\documentclass[review]{elsarticle}

\usepackage{lineno,hyperref}
\usepackage{bm}
\usepackage{amsmath}
\usepackage{amsfonts}       
\usepackage{booktabs}       
\usepackage{algorithm}
\usepackage{algpseudocode}
\usepackage{subfigure}
\usepackage{multirow}
\usepackage{threeparttable}
\usepackage{color}
\graphicspath{{figures/}}
\modulolinenumbers[5]

\journal{Knowledge-Based Systems}

\begin{document}

\begin{frontmatter}

\title{Adversarially Regularized Graph Attention Networks for Inductive Learning on Partially Labeled Graphs}

\author[mymainaddress]{Jiaren Xiao}
\author[mysecondaryaddress]{Quanyu Dai}
\author[mythirdaddress,myfourthaddress]{Xiaochen Xie}
\author[mymainaddress]{James Lam}
\author[mymainaddress]{Ka-Wai Kwok\corref{mycorrespondingauthor}}
\cortext[mycorrespondingauthor]{Corresponding author}
\ead{kwokkw@hku.hk}

\address[mymainaddress]{Department of Mechanical Engineering, The University of Hong Kong, Hong Kong, China}
\address[mysecondaryaddress]{Department of Computing, The Hong Kong Polytechnic University, Hong Kong, China}
\address[mythirdaddress]{Department of Automation, Harbin Institute of Technology, Shenzhen, China}
\address[myfourthaddress]{Guangdong Key Laboratory of Intelligent Morphing Mechanisms and Adaptive Robotics, Shenzhen, China}

\begin{abstract}
The high cost of data labeling often results in node label shortage in real applications. To improve node classification accuracy, graph-based semi-supervised learning leverages the ample unlabeled nodes to train together with the scarce available labeled nodes. However, most existing methods require the information of all nodes, including those to be predicted, during model training, which is not practical for dynamic graphs with newly added nodes. To address this issue, an adversarially regularized graph attention model is proposed to classify newly added nodes in a partially labeled graph. An attention-based aggregator is designed to generate the  representation of a node by aggregating information from its neighboring nodes, thus naturally generalizing to previously unseen nodes. In addition, adversarial training is employed to improve the model's robustness and generalization ability by enforcing node representations to match a prior distribution. Experiments on real-world datasets demonstrate the effectiveness of the proposed method in comparison with the state-of-the-art methods. The code is available at \url{https://github.com/JiarenX/AGAIN}.
\end{abstract}

\end{frontmatter}


\section{Introduction}
\label{intro}
Graphs naturally represent the data with complicated relationships and rich information, as seen in social, biological, and citation networks. Since graph-structured data are usually sparse, nonlinear and high-dimensional, the analysis of graph-structured data is challenging. To tackle graph-analytic tasks, a common approach  is graph embedding which aims at learning the low-dimensional node representation vectors~\cite{goyal_graph_2018}. The key idea is to encode meaningful information like node features and graph structure into node representations (i.e., embedding vectors). Based on graph embedding, node classification tasks can be performed using classical machine learning techniques like a linear support vector machine (SVM) classifier~\cite{cui_survey_2018}. Node classification has many practical applications, such as predicting user types in e-commence networks \cite{eswaran_zoobp_2017}, assigning topics to papers in citation networks~\cite{kipf_semi-supervised_2017}, and classifying protein roles in biological networks \cite{hamilton_inductive_2017}. However, in many scenarios, labels are only available for a small subset of nodes due to the high cost and technical difficulty of labeling by human. To lessen the requirement of large amounts of labeled training nodes, a recent surge of research interest can be seen in semi-supervised learning on graphs~\cite{kipf_semi-supervised_2017}. 

Graph-based semi-supervised learning leverages the ample unlabeled nodes to train together with very few labeled nodes, so that the node classification accuracy can be improved. In recent years, substantial research effort has been devoted to designing neural network models that directly operate on graphs, known as graph neural networks (GNNs)~\cite{kipf_semi-supervised_2017, wu_comprehensive_2020}. In addition, a few GNN models introduce attention mechanisms to graph embedding~\cite{lee_attention_2019, velickovic_graph_2018}. Attention mechanisms allow the graph embedding model to highlight neighbors that contain more task-relevant information, consequently increasing the model capacity. Moreover, attention mechanisms are likely to help the model to disregard the noisy portions of a graph, thus improving model robustness.

Most existing graph embedding approaches are inherently transductive, since they focus on generating representations for nodes in a fixed graph. These methods optimize the representation of each node based on random walk \cite{perozzi_deepwalk_2014, tang_line_2015} or matrix factorization \cite{cao_grarep_2015, ou_asymmetric_2016}. To predict nodes newly added to the graph (i.e., unseen nodes), the transductive methods need non-trivial modifications and additional training, consequently suffering from expensive computation~\cite{hamilton_inductive_2017}. Many real applications involve unseen nodes. For example, new members may join social networks in Twitter and Facebook. In addition, there are usually massive amounts of new publications added to citation networks in databases like PubMed and arXiv. An inductive approach~\cite{hamilton_inductive_2017, yang_revisiting_2016}, that enables node representations to be quickly generated for unseen nodes, is essential for such scenarios. Compared to transductive learning, the inductive learning problem is more challenging, since the model would have already been optimized on the existing nodes before the introduction of new nodes.

Furthermore, noise, perturbations or even attacks are commonly seen in graph-structured data. For instance, scientific papers may have spelling mistakes, missing words, or incorrect expressions; criminals tend to hide or fabricate their personal information in social networks; fraudsters often manipulate the online reviews of their products to attract customers on e-commerce platforms. The general learning objective of existing graph embedding methods is to well preserve graph structure only, or to jointly capture both structural properties and side information like node features. As a result, the noise in structure and features can lead to poor performance of these methods \cite{bojchevski_adversarial_2019,dai_adversarial_2018-1, zugner_adversarial_2018}. In semi-supervised learning, a common regularization is to drive connected nodes to have the same label based on the homophily assumption~\cite{mcpherson_birds_2001, yang_revisiting_2016}. As shown in \cite{li_deeper_2018}, the working mechanism of graph convolution \cite{kipf_semi-supervised_2017} is a special form of Laplacian smoothing which mixes the features of a node and its neighbors. Therefore, the relational effect of graph structures \cite{zugner_adversarial_2018} is likely to worsen the model performance, since manipulating one node or edge may misguide the predictions of relational nodes. 

To improve the model robustness over noisy graphs, some pioneering research \cite{dai_adversarial_2018, pan_adversarially_2018, yu_learning_2018} employs adversarial training in graph embedding. These studies are largely inspired by recent generative adversarial models \cite{donahue_adversarial_2017, glover_modeling_2016, radford_unsupervised_2016}, which are shown to be effective in learning robust representations. Similar to the adversarial autoencoder (AAE)~\cite{makhzani_adversarial_2016}, the basic idea is to match the learned node representations with a prior distribution using adversarial training. The underlying motivation is to enforce an additional regularization on the node representations, and to introduce a certain amount of uncertainty in the learning process. This helps improve the model robustness against noisy graphs. Adversarial training also upholds the potential to avoid overfitting and achieve relatively promising generalization performance. However, to our knowledge, none of these prior studies focuses on robust graph embedding under the inductive semi-supervised setting.

In this paper, we propose a novel method named $ \underline{\rm A} $dversarially regularized $ \underline{\rm G} $raph $ \underline{\rm A} $ttention networks for $ \underline{\rm IN} $ductive learning on partially labeled graphs (AGAIN). On one hand, our method encodes graph structure and node features into node embeddings with an attention-based aggregator. When aggregating the neighborhood information, an attention mechanism is adopted to assign different learnable weights to the sampled neighbors, capturing the importance of each neighbor. At the inference time, the learned aggregator can produce informative representations for previously unseen nodes. On the other hand, adversarial training is employed to learn robust node representations by enforcing the representations to match a prior distribution.

The proposed method is evaluated on four datasets including three citation networks (i.e., Cora, CiteSeer and PubMed) as well as one social network named BlogCatalog. The \textit{main contributions} of this work are summarized as follows.
\begin{itemize}
	\item The first adversarially regularized GNN model is proposed and designed specifically to address the challenging inductive learning problem on partially labeled graphs.
    \item Our model is devised to incorporate attention mechanism and adversarial training, effectively generating informative and robust node representations.	
	\item Extensive experiments are conducted with real-world information networks, showing our model is comparable with or even superior to the state-of-the-art methods on the benchmark inductive node classification tasks.	
\end{itemize}

This rest of this paper is organized as follows. The relevant literature is reviewed in Section~\ref{rel_work}. The proposed method is described in Section~\ref{pro_meth}. The experimental results are reported in Section~\ref{exp}. Finally, the conclusions are summarized in Section~\ref{con}.
\section{Related Work}
\label{rel_work}
\subsection{Graph-based Semi-supervised Learning}
On a partially labeled graph, graph-based semi-supervised learning aims to jointly utilize both the scarce labeled and ample unlabeled nodes to improve node classification accuracy. There exist two learning paradigms: transductive learning and inductive learning. Transductive learning \cite{zhou_learning_2004, zhu_semi-supervised_2003} only aims at classifying the unlabeled nodes that are observed in training time. Inductive learning algorithms, such as manifold regularization \cite{belkin_manifold_2006} and semi-supervised embedding \cite{weston_deep_2008}, can generalize to unobserved nodes. Planetoid \cite{yang_revisiting_2016} has both transductive and inductive variants. The inductive algorithm, Planetoid-I, learns a parameterized classifier based on node features to facilitate predictions on nodes unseen during training. Note that, graph-based semi-supervised learning assumes the training and test nodes share the same label space. In contrast, open-set learning~\cite{fang_learning_2021} and out-of-distribution detection~\cite{fang_is_2022} have test data from the classes that are unseen in training data.

Graph embedding is a broader research topic that focuses on mapping the nodes to representation vectors in the low-dimensional space. There are a number of recent approaches that learn low-dimensional embeddings based on random walk (e.g., DeepWalk \cite{perozzi_deepwalk_2014}, LINE \cite{tang_line_2015}, and node2vec \cite{grover_node2vec_2016}) and matrix factorization (e.g., GraRep \cite{cao_grarep_2015}, HOPE \cite{ou_asymmetric_2016}, and M-NMF \cite{wang_community_2017}). The learning objective of these methods is to maximally preserve the topological information. Under the assumption that node features are available, some approaches are capable of exploiting both the topological and feature information, such as TADW \cite{yang_network_2015}, TriDNR \cite{pan_tri-party_2016}, and UPP-SNE \cite{zhang_user_2017}. However, these methods are transductive by training embeddings for individual nodes in a fixed graph, and not designed specifically for semi-supervised learning.

Beyond the classical graph embedding methods, increasing research interest can be seen in graph neural networks (GNNs)~\cite{wu_comprehensive_2020} which can be categorized as spectral and spatial approaches. Spectral-based approaches introduce filters for graph convolutions \cite{bruna_spectral_2014, defferrard_convolutional_2016, kipf_semi-supervised_2017}. Among them, Kipf and Welling \cite{kipf_semi-supervised_2017} simplified the previous spectral convolutions to be a localized first-order approximation for semi-supervised learning. This algorithm depends on the graph Laplacian and all node features during training, and hence lies within the transductive setting. Imitating the convolutional neural networks on images, the spatial approaches define graph convolution directly based on the spatial relations of a node and its neighborhood \cite{gilmer_neural_2017, hamilton_inductive_2017, velickovic_graph_2018}. The well-known inductive method, GraphSAGE \cite{hamilton_inductive_2017}, computes the node embeddings by sampling a fixed-size neighborhood and then aggregating features. The feature aggregation is based on the elementwise mean of neighborhood (GS-mean), the inductive variant of GCN \cite{kipf_semi-supervised_2017} (GS-GCN), the LSTM architecture (GS-LSTM), and elementwise max-pooling or mean-pooling operation (GS-pool). The performance of GraphSAGE in several large-scale benchmarks is quite impressive. 

Furthermore, attention mechanisms have been widely adopted in computer vision~\cite{mnih_recurrent_2014} and natural language processing~\cite{bahdanau_neural_2015}. The goal is to attend over important parts of the data, and to improve the performance of a machine learning model. Attention mechanisms have also been introduced to designing GNN models. GAT~\cite{velickovic_graph_2018} assigns learnable weights to the entire neighborhood nodes, yielding improved or matched performance in semi-supervised node classification. Although the reported single-graph experiments are transductive, GAT is capable of supporting inductive learning on one graph. The reason is that GAT only requires access to the local neighborhood of a node, instead of the upfront knowledge about the whole graph. In addition, GAT adds a self-loop to a node and treats the node itself as one of its neighbors, so that the previous node representation can be inherently incorporated in the neighborhood aggregation process. Unlike the traditional multi-head attention, GaAN~\cite{zhang_gaan_2018} controls the importance of each attention head with a convolutional subnetwork. HAN \cite{wang_heterogeneous_2019} proposes a two-level attention (i.e., node level and semantic level) for learning on heterogeneous graphs. 

Attention mechanism is also explored in this work. However, different from GAT, we sample a fixed size of neighboring nodes before calculating attention coefficients, in order to keep the computational footprint consistent for every node. Additionally, we utilize a skip connection~\cite{he_identity_2016} to incorporate the node representation of the previous layer. As introduced in GraphSAGE~\cite{hamilton_inductive_2017}, such skip connection operation has the potential to boost model performance. Moreover, the methods introduced above are mostly unregularized and ignore the data distribution of learned node representations, which may result in poor performance on sparse and noisy graphs in real applications. In this work, we utilize adversarial training to address this issue.
\subsection{Graph Adversarial Attacks and Defenses}
Many studies on image \cite{bhagoji_role_2020, goodfellow_explaining_2015, zhang_principal_2020} and text \cite{jia_adversarial_2017} have shown that neural networks are vulnerable to deliberate adversarial perturbations in the input. There are two dominant types of adversarial attacks \cite{biggio_security_2014, papernot_limitations_2016}, namely, poisoning attacks in which the model is trained after the attack, and evasion attacks targeting the test phase in which the learned model is assumed to be fixed. Recently, it is also found that the performance of graph embedding methods including GNNs  would drop significantly under malicious manipulations in graph structure or node features \cite{bojchevski_adversarial_2019, dai_adversarial_2018-1, zugner_adversarial_2018}. Accordingly, some defense models are proposed to improve the robustness of GNNs \cite{wu_graph_2020, zugner_certifiable_2019}. An additional hinge loss is considered in \cite{zugner_certifiable_2019} during the training process to achieve certified robustness under perturbations on the node features. Inherited from the principle of information bottleneck \cite{tishby_information_2000, tishby_deep_2015}, GIB \cite{wu_graph_2020} learns minimal sufficient node representations that naturally defend against attacks. To evaluate the model robustness, GIB employs adversarial attacks generated using Nettack \cite{zugner_adversarial_2018}, and simple feature attacks which inject Gaussian noise into the feature vectors. In this work, similar feature attacks are also used for robustness evaluation, due to the generality of Gaussian noise injection.

\subsection{Generative Adversarial Models}
The deep generative model, i.e., generative adversarial networks (GANs) \cite{goodfellow_generative_2014}, builds a minimax adversarial game for two players: the generator and the discriminator. The discriminator is usually a multi-layer perceptron (MLP) which is trained to tell apart whether an input sample comes from the real data distribution or the generator. Simultaneously, the generator is trained to generate samples as close to the real samples as possible to fool the discriminator. Being inspired by GANs, Makhzani et al. \cite{makhzani_adversarial_2016} employed adversarial training to perform variational inference by matching the representations with a prior distribution. This adversarial autoencoder (AAE) achieves competitive performance in semi-supervised classification on images. Some other generative adversarial models are proposed to learn robust representations for images \cite{donahue_adversarial_2017, radford_unsupervised_2016} and text \cite{glover_modeling_2016}. 

Recently, the adversarial regularization has been applied to graph-structured data in several studies. The first one is ANE \cite{dai_adversarial_2018} which combines an inductive variant of DeepWalk and the adversarial training for learning robust node representations. ARGA~\cite{pan_adversarially_2018} and ARVGA \cite{pan_adversarially_2018} further utilize the node features together with topological information in a similar adversarial learning scheme. NetRA \cite{yu_learning_2018} circumvents the need of a pre-defined fixed prior, and further employs Wasserstein GANs \cite{arjovsky_wasserstein_2017} to overcome the unstable problem during training. However, the inductive semi-supervised learning that this work focuses on is not considered in the prior art.
\section{Proposed Method}
\label{pro_meth}
In this section, we first introduce the problem and main notations. Then we present an overview of the model architecture, followed by a detailed description of each component. Finally, the algorithm of our model is provided together with an analysis of the computational complexity.
\subsection{Problem Definition and Notations}
\label{pro_defi}
An information network can be expressed as an attributed graph $ \mathcal{G}\left( \bm{V},\bm{E},\bm{X}\right) $, where $ \bm{V} $ is the set of nodes, $ \bm{E} $ is the set of edges representing the relationships between nodes, and $ \bm{X}\in \mathbb{R}^{N\times D} $ is the feature matrix. $ N $ is the number of nodes and $ D $ is the feature dimension. $ \bm{x}_v^{\top} $ is one row in the feature matrix $ \bm{X} $ representing the feature vector of node $ v\in \bm{V} $. The topological structure of unweighted graph $ \mathcal{G} $ can be represented as an adjacency matrix $ \bm{A}\in \mathbb{R}^{N\times N} $ with each element $ a_{ij} $ set as 0 or 1, specifying whether an edge exists between two nodes. The degree of the $ i $-th node $ v $ is the number of its connected edges, i.e., $ {\rm degree}\left( v\right)  = \sum\nolimits_{j} a_{ij} $. The average degree is further defined as $ \left\langle k \right\rangle = 2\left| \bm{E} \right|/N $, indicating the density of an undirected graph.

\begin{figure}[tbp]
	\centering
	\includegraphics[width=12.05cm]{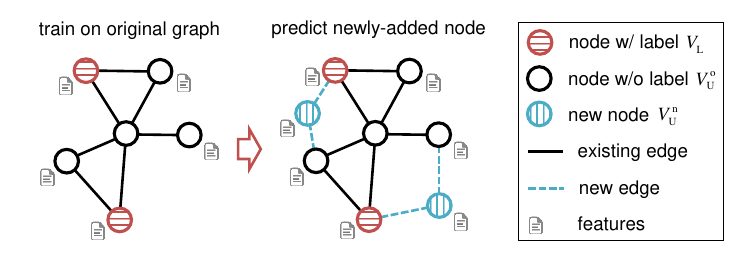}
	\caption{Inductive learning under semi-supervised setting. The node classification model is trained on the original graph in which only a small percentage of nodes have labels. Then the learned model is directly applied to make predictions on nodes that are newly added and unseen during training.}
	\label{fig_3_1}
\end{figure}
\begin{table}[htbp]
	\caption{Main notations.}
	\label{tab_0}
	\centering
	\begin{tabular}{ll}
		\toprule
		Notation                                                                                               & Description                                                                                           \\ \midrule
		$ \mathcal{G} $                                                                                        & An attributed graph                                                                                   \\
		$ \bm{V} $, $ \bm{E} $, $ \bm{A} $                                                                                             & Node set, edge set, and binary adjacency matrix of $ \mathcal{G} $                                                                                                                                     \\
		$ \bm{x}_v $, $ \bm{X} $                                                                                           & Feature vector of node $ v\in \bm{V} $ and feature matrix of $ \mathcal{G} $                                                                                                                                 \\
		$ \bm{u}_{v} $, 		$ \bm{U} $                                                                                             & Embedding vector of node $ v\in \bm{V} $ and  representation matrix of $ \mathcal{G} $                                                                                                                         \\
		$ \bm{Y} $, $ \hat{\bm{Y}} $                                                                                             & Label matrix and prediction score matrix of $ \mathcal{G} $                                                      \\ \midrule
		$ N $                                                                                                  & Number of nodes in $ \mathcal{G} $                                                                    \\
		$ n $                                                                                                  & Number of labeled nodes per class in $ \mathcal{G} $                                                  \\
		$ \left| \bm{E} \right| $, 		$ \left\langle k \right\rangle $                                                                              & Number of edges and average degree in $ \mathcal{G} $                                                                                                                                        \\
		$ D $, $ d $                                                                                                  & Feature dimension and embedding dimension                                                                                                                                                                      \\
		$ C $                                                                                                  & Number of classes in $ \bm{Y} $                                                                       \\ \midrule
		$ f_{\bm{\varphi}}\left( \cdot\right)  $, $ l_{\bm{\psi}}\left( \cdot\right)  $, $ d_{\bm{w}}(\cdot) $ & GNN encoder, node classifier, and discriminator                                            \\
		$ \bm{\varphi}, \bm{\psi}, \bm{w} $                                                                    & Sets of parameters in $ f_{\bm{\varphi}}(\cdot) $, $ l_{\bm{\psi}}(\cdot) $ and $ d_{\bm{w}}(\cdot) $ \\
		$ n_{\rm max} $                                                                                        & Maximum training epoch                                                                                \\
		$ n_{\rm D} $                                                                                          & Number of discriminator training per generator iteration                                                        \\
		$ K $                                                                                                  & Maximum search depth                                                                                  \\
		$ \bm{B} $ & A batch of nodes
		\\
		$ p_{r} $ & Discriminator learning rate
		\\
		$ p_{c} $ & Weight decay coefficient
		\\
		$ s $                                                                                                  & Neighborhood sample size
		\\
		$ \sigma $                                                                                             & Nonlinear activation function                                                                                                                                          \\
		$ {\rm AGG} $                                                                                          & Aggregator function                                                                                                                                                           
        \\
        $ \alpha $ & Attention coefficient
        \\
        $ \bm{h} $ & Latent representation		
		\\ \midrule
		$ \eta $                                                                                             & Percentage of nodes with noise                                                                        \\
		$ \lambda $                                                                                            & Feature noise ratio                                                                                   \\
		$ \Delta $                                                                                             & Performance gap                                                                                       \\ \bottomrule
	\end{tabular}
\end{table}

As shown in \textbf{Figure~\ref{fig_3_1}}, in this work, we investigate the classification of nodes that are newly added to a partially labeled attributed graph. The main notations used in this paper are summarized in \textbf{Table~\ref{tab_0}}. A set of nodes, $ \bm{V} $, consists of labeled nodes $ \bm{V}_{\rm L} $ and unlabeled nodes $ \bm{V}_{\rm U} $. Some of the unlabeled nodes (i.e., $ \bm{V}_{\rm U}^{\rm o} $) are observed during training, and the rest (i.e., $ \bm{V}_{\rm U}^{\rm n} $) are unobserved. Unobserved nodes, $ \bm{V}_{\rm U}^{\rm n} $, are added to the original graph during test phase.

Graph embedding aims at mapping a node, $ v\in \bm{V} $, to a low-dimensional embedding vector $ \bm{u}_{v} $. $ \bm{u}_{v}^{\top} $ is one row within representation matrix $ \bm{U}\in \mathbb{R}^{N\times d} $, where $ d $ is the embedding dimension. As shown in \textbf{Figure~\ref{fig_3_1}}, the attributed graph is partially labeled, that is, only a small percentage of nodes are with labels. To perform node classification on top of embeddings, the semi-supervised learning is defined as learning a classifier, $ f:\bm{V}\longmapsto \bm{Y} $, using both labeled nodes (i.e., $ \bm{V}_{\rm L} $) and observed unlabeled nodes (i.e., $ \bm{V}_{\rm U}^{\rm o} $). Label matrix, $ \bm{Y}\in\mathbb{R}^{N\times C} $, contains binary element, $ Y_{vk} $, indicating whether node $ v $ is associated with class $ k $. The total number of classes in $ \bm{Y} $ is $ C $. There are two learning paradigms. The transductive learning only aims to predict the observed unlabeled nodes in the graph, that is, $ \bm{V}_{\rm U}^{\rm o} $. Inductive learning further seeks to generalize the classification model to nodes that are unseen in the graph during training, that is, $ \bm{V}_{\rm U}^{\rm n} $. This work focuses on the inductive semi-supervised learning.

As stated in Section~\ref{intro}, the robustness of a graph embedding model against noisy input is an important issue, since noise and perturbations are commonly seen in graph-structured data. Therefore, in this work, we assume the inputs to be noisy when evaluating robustness. For an attributed graph, these inputs are usually feature matrix and structural information such as adjacency matrix, PPMI matrix \cite{levy_neural_2014} and random walk. Being inspired by the feature attacks in GIB \cite{wu_graph_2020}, we randomly select a percentage of nodes in the graph, and add independent Gaussian noise to each dimension of the node features. The Gaussian noise is injected during the test phase in which the model is fixed. As introduced in Section~\ref{rel_work}, this kind of noise injection belongs to the evasion attacks. As shown in \cite{wu_graph_2020}, the resilience to feature attacks, or the lack of it, can be reflected by the consequent performance under feature noise.

\subsection{Overview of Model Architecture}
\begin{figure}[tbp]
	\centering
	\includegraphics[width=10.1cm]{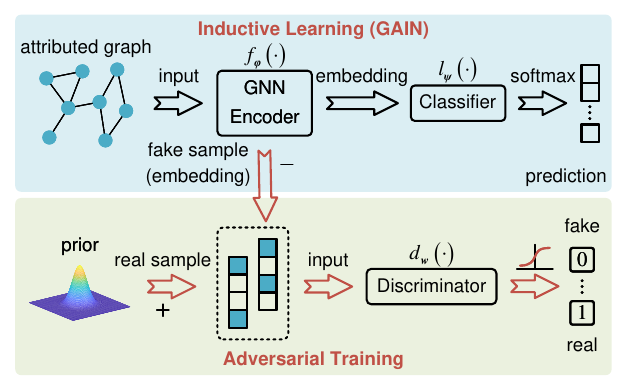}
	\caption{Model architecture of AGAIN. The upper and lower tiers illustrate inductive learning and adversarial training, respectively. GNN encoder is empowered by the attention-based aggregator.}
	\label{fig_1}
\end{figure}

\textbf{Figure~\ref{fig_1}} shows the model architecture of the proposed method, i.e., AGAIN. There are two main components, i.e., inductive learning and adversarial training. Specifically, the $ \underline{\rm G} $raph $ \underline{\rm A} $ttention networks for $ \underline{\rm IN} $ductive learning (GAIN) consist of GNN encoder $ f_{\bm{\varphi}}(\cdot) $ and node classifier $ l_{\bm{\psi}}(\cdot) $. GNN encoder encodes the topological information and node features of an input graph into low-dimensional node embedding vectors with the attention-based aggregator. Node embeddings are further transformed by a node classifier, $ l_{\bm{\psi}}(\cdot) $, which is a fully-connected layer followed by a softmax activation, into predictions of node labels. Moreover, the adversarial training imposes a prior distribution on the node embeddings. Discriminator, $ d_{\bm{w}}(\cdot) $, aims at discriminating the prior samples and the  embedding vectors. It is a standard multi-layer perceptron (i.e., MLP), in which the output is a single neuron followed by a sigmoid activation, indicating the probability of an input sample to be real. Note that GNN encoder $ f_{\bm{\varphi}}(\cdot) $ also plays the role of generating fake samples (i.e., embedding vectors) in the adversarial training. Hence, the GNN encoder is shared by the inductive learning and adversarial training components. Three sets of parameters, $ \bm{\varphi}, \bm{\psi}$ and $ \bm{w} $, are involved in $ f_{\bm{\varphi}}(\cdot) $, $ l_{\bm{\psi}}(\cdot) $ and $ d_{\bm{w}}(\cdot) $, respectively.
 
\subsection{Inductive Learning}
\label{gat}
\begin{figure}[htbp]
	\centering
	\includegraphics[width=10.97cm]{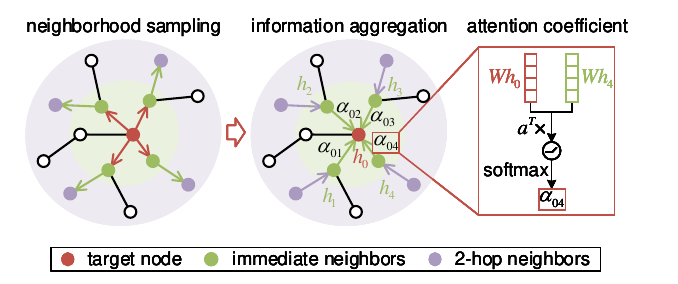}
	\caption{Illustration of the neighborhood sampling and the subsequent information aggregation process. The sign “$ \times $” denotes matrix multiplication. Attention coefficient is activated by the leaky ReLU nonlinearity before softmax operation.}
	\label{fig_3.3_1}
\end{figure}

As illustrated in \textbf{Figure~\ref{fig_3.3_1}}, in the neighborhood sampling stage, rather than considering the whole neighborhood of a given target node, a fixed size of neighbors are randomly sampled at each search depth. In case that sample size is larger than the node degree, neighbors are sampled with replacement. The sampling is an outward process in which the far neighborhood is gradually discovered. The maximum search depth is denoted as $ K $. Then the nodes aggregate information from their sampled neighbors. Note that the aggregation is an inward process. As the process iterates, more and more information is gained from far neighborhood by the target node. 

When aggregating neighborhood information, we introduce an attention mechanism \cite{velickovic_graph_2018} to assign different learnable weights to the neighbors, indicating their relative importance in assisting the learning of target node. As shown in \textbf{Figure~\ref{fig_3.3_1}}, at step $ k $, attention coefficient $ \alpha_{vu}^{k} $ can be computed as follows.
\begin{equation}
\alpha_{vu}^{k}=\dfrac{{\rm exp}\left(\sigma_{1}\left(\left(\bm{a}^{k}\right)^{\top}[\bm{W}^{k}\bm{h}_{v}^{k-1}; \bm{W}^{k}\bm{h}_{u}^{k-1}]\right)\right)}{\sum\nolimits_{m\in \bm{S}_{v}} {\rm exp}\left(\sigma_{1}\left( \left(\bm{a}^{k}\right)^{\top}[\bm{W}^{k}\bm{h}_{v}^{k-1}; \bm{W}^{k}\bm{h}_{m}^{k-1}]\right)\right)}
\label{eq_-1}
\end{equation}
where $ \bm{S}_{v} $ is the set of immediate neighbors of node $ v $; $ \bm{h}_{v}^{k-1} (v\in \bm{V}) $ and $ \bm{h}_{u}^{k-1} (u\in \bm{S}_{v}) $ are the latent representations of target node $ v $ and neighboring node $ u $ at the previous step (i.e., $ k-1 $), respectively; $ \bm{a}^{k} $ and $ \bm{W}^{k} $ are the weight vector and matrix for linear transformations, respectively. Note that, at step $ k=0 $, the latent representation is the node feature vector, that is, $ \bm{h}_{v}^{0}=\bm{x}_{v} $. Therefore, the latent representations are initialized with node features and updated step by step. Here, nonlinear activation, $ \sigma_{1} $, is a leaky ReLU function, i.e., $ \sigma_{1}(x)={\rm max}(0.2x,x) $. 

The latent representation of neighborhood can then be derived as follows.
\begin{equation}
\bm{h}_{S}^{k}={\rm AGG}_{k}(\bm{h}_{u}^{k-1}\mid u\in \bm{S}_{v}) =\sum\nolimits_{u\in \bm{S}_{v}}\alpha_{vu}^{k}\bm{h}_{u}^{k-1}
\label{eq_0}
\end{equation}
in which $ {\rm AGG}_{k} $ is the aggregator function at step $ k $. Then the latent representation of node $ v $ at step $ k $ (i.e., $ \bm{h}_{v}^{k} $) can be calculated. 
\begin{equation}
\bm{h}_{v}^{k}=\sigma_{2}({[\bm{W}_{v}^{k}\bm{h}_{v}^{k-1}; \bm{W}_{S}^{k}\bm{h}_{S}^{k}]}),
\label{eq_emb_k_1}
\end{equation} 
\begin{equation}
\bm{h}_{v}^{k}=\bm{h}_{v}^{k}/\parallel \bm{h}_{v}^{k} \parallel_{2},
\label{eq_emb_k_2}
\end{equation}
where $ \bm{W}_{v}^{k} $ and $ \bm{W}_{S}^{k} $ are also weight matrices for linear transformations; nonlinear activation, $ \sigma_{2}(x)={\rm max}(0,x) $, is a ReLU function. Note that we implement a skip connection~\cite{he_identity_2016} in Eq.~\ref{eq_emb_k_1} to incorporate the node representation of previous layer. As introduced in GraphSAGE~\cite{hamilton_inductive_2017}, such skip connection operation can potentially boost model performance.

The final representation output at step $ K $ is denoted as $ \bm{u}_{v} $, which is the learned representation (i.e., embedding vector) of node $ v $. 
\begin{equation}
\bm{u}_{v}=\bm{h}_{v}^{K}=f_{\bm{\varphi}}\left(\bm{x}_{v}, \bm{x}_{S}\right) , v\in \bm{V},
\label{eq_emb}
\end{equation}
in which $ f_{\bm{\varphi}} $ is the GNN encoder, $ \bm{x}_{v} $ is the feature vector of node $ v $, and $ \bm{x}_{S} $ is the feature matrix of the sampled neighboring nodes. For notational convenience, in the following descriptions, we simply use $ f_{\bm{\varphi}}(\bm{x}_{v}) $ to denote $ \bm{u}_{v} $. Note that embedding vector, $ \bm{u}_{v} $, is also the fake sample in adversarial training indicated by a sign “$ - $” in \textbf{Figure~\ref{fig_1}}.

Finally, prediction score vector, $ \hat{\bm{y}}_{v} $, can be calculated by feeding embedding vector $ \bm{u}_{v} $ into node classifier $ l_{\bm{\psi}}(\cdot) $.
\begin{equation}
\hat{\bm{y}}_{v}=l_{\bm{\psi}}(\bm{u}_{v}), v\in \bm{V}.
\label{eq_0.5}
\end{equation}
$ \hat{\bm{y}}_{v}^{\top} $ is one row in prediction score matrix $ \hat{\bm{Y}}\in\mathbb{R}^{N\times C} $. Under semi-supervised setting, the inductive learning component is trained by minimizing the cross-entropy loss of labeled nodes as follows. 
\begin{equation}
\mathcal{L}_{\rm GAIN}=-\mathop{\mathbb{E}}\limits_{v\in \bm{B}}\left[ \sum_{k=1}^C Y_{vk}{\rm log}(\hat{Y}_{vk})\right] 
\label{eq_1}
\end{equation}
where $ \bm{B} $ is a sampled batch from the training nodes; binary element $ Y_{vk} $ within label matrix $ \bm{Y} $ indicates whether a node $ v\in \bm{B} $ belongs to class $ k $; and $ \hat{Y}_{vk} $ is the corresponding element in prediction score matrix $ \hat{\bm{Y}} $. 

\subsection{Adversarial Training}
An adversarial training model is employed to regularize the embedding vectors. The learned embeddings can be enforced to match a certain prior distribution. It builds an adversarial training platform for two players, namely, generator $ g_{\bm{\theta}}(\cdot) $ and discriminator $ d_{\bm{w}}(\cdot) $, to play a minimax game. Specifically, generator, $ g_{\bm{\theta}}(\cdot) $, represents a nonlinear transformation from the input graph to embedding vectors. In this work, GNN encoder, $ f_{\bm{\varphi}}(\cdot) $, play the role of $ g_{\bm{\theta}}(\cdot) $. A real sample, $ \bm{z} $, is sampled from prior distribution $ P_{\rm g}(\bm{z}) $, while embedding vector $ f_{\bm{\varphi}}(\bm{x}) $ is treated as the fake sample. Discriminator, $ d_{\bm{w}}(\cdot) $, is a standard multi-layer perceptron. The output of discriminator, which is of one dimension followed by a sigmoid activation, indicates the probability of an input sample to be real. The value function of adversarial training can be expressed as follows \cite{goodfellow_generative_2014}. 
\begin{equation}
\mathop{\rm min}\limits_{\bm{\varphi}} \mathop{\rm max}\limits_{\bm{w}} \mathop{\mathbb{E}}\limits_{\bm{z}\sim P_{\rm g}(\bm{z})}[{\rm log}d_{\bm{w}}(\bm{z})] + \mathop{\mathbb{E}}\limits_{\bm{x}\sim P_{\rm data}(\bm{x})}[{\rm log}(1-d_{\bm{w}}(f_{\bm{\varphi}}(\bm{x})))]
\label{eq_2}
\end{equation}
in which $ P_{\rm data}(\bm{x}) $ is the feature distribution of nodes.

During training, the discriminator is trained to distinguish prior samples from embedding vectors, while the generator aims to fit node embeddings to the prior distribution, thus misguiding the discriminator. We can separate the training of discriminator and generator. The loss function of discriminator is defined as
\begin{equation}
\mathcal{L}_{\rm DIS}({\bm{w}};\bm{x},\bm{z}) = -\mathop{\mathbb{E}}\limits_{\bm{z}\sim P_{\rm g}(\bm{z})}[{\rm log}d_{\bm{w}}(\bm{z})] - \mathop{\mathbb{E}}\limits_{\bm{x}\sim P_{\rm data}(\bm{x})}[{\rm log}(1-d_{\bm{w}}(f_{\bm{\varphi}}(\bm{x})))].
\label{eq_3}
\end{equation}
The loss function of generator is
\begin{equation}
\mathcal{L}_{\rm GEN}({\bm{\varphi}};\bm{x}) = - \mathop{\mathbb{E}}\limits_{\bm{x}\sim P_{\rm data}(\bm{x})}[{\rm log}(d_{\bm{w}}(f_{\bm{\varphi}}(\bm{x})))].
\label{eq_4}
\end{equation}
In many practices of previous studies \cite{donahue_adversarial_2017, makhzani_adversarial_2016}, the Gaussian or Uniform distribution is chosen as a prior for learning robust representations. Note that, in this work, the prior distribution produces real samples, rather than serving as a source of noise for generating fake samples as in \cite{goodfellow_generative_2014}. 
\subsection{Algorithm}
\subsubsection{AGAIN Minibatch Training}
\begin{algorithm}[htbp]
	\caption{AGAIN Minibatch Training}
	\label{alg_1}
	\begin{algorithmic}[1]
		\Require Graph $ \mathcal{G}\left( \bm{V},\bm{E},\bm{X}\right) $; maximum training epoch $ n_{\rm max} $; maximum search depth $ K $; number of discriminator training per generator iteration $ n_{\rm D} $; attention-based aggregator function $ {\rm AGG}_{k} $ (including weight vector $ \bm{a}^{k} $, weight matrix $ \bm{W}^{k} $, and nonlinear activation $ \sigma_{1} $), $ k\in\left\lbrace 1,\ldots,K \right\rbrace$; weight matrices $ \bm{W}_{v}^{k} $ and $ \bm{W}_{S}^{k} $, $ k\in\left\lbrace 1,\ldots,K \right\rbrace$; nonlinearity $ \sigma_{2} $.
		
		\For{$ {\rm epoch}<n_{\rm max} $}
		\State Sample a batch of labeled nodes (i.e., $ \bm{B} $) with initial representations set as $ \bm{h}_{v}^{0}=\bm{x}_{v} (v\in \bm{B}) $ and sample the neighboring features $ \bm{x}_{S} $ (including those of the immediate neighbors, i.e., $ \bm{S}_{v} $).
		\label{emb_start}		 
		\For {$ k=1,\ldots,K $}		 
		\State $ \bm{h}_{S}^{k}={\rm AGG}_{k}(\bm{h}_{u}^{k-1}\mid u\in \bm{S}_{v}) $
		\State $ \bm{h}_{v}^{k}=\sigma_{2}({[\bm{W}_{v}^{k}\bm{h}_{v}^{k-1}; \bm{W}_{S}^{k}\bm{h}_{S}^{k}]}) $
		\State $ \bm{h}_{v}^{k}=\bm{h}_{v}^{k}/\parallel \bm{h}_{v}^{k} \parallel_{2}$
		\EndFor
		\State $ \bm{u}_{v}=\bm{h}_{v}^{K}=f_{\bm{\varphi}}(\bm{x}_{v}, \bm{x}_{S}) $, $ \hat{\bm{y}}_{v}=l_{\bm{\psi}}(\bm{u}_{v}) $.
		\State Compute the cross-entropy loss using Eq.~\ref{eq_1}
		\State Backpropagate loss and update $ {\bm{\varphi}} $ and $ {\bm{\psi}} $
		\label{emb_end}
		\For{$ n<n_{\rm D} $}
		\label{disc_start}
		\State Sample a batch of nodes $  \bm{x}_{v} (v\in \bm{B}) $ and compute embeddings $ \bm{u}_{v} $
		\State Sample a batch from the prior distribution $ \bm{z}_{i}\sim P_{\rm g}(\bm{z}) (i=1,\ldots, \left| \bm{B} \right|)$ 
		\State Compute $ \mathcal{L}_{\rm DIS} $ using Eq.~\ref{eq_3}
		\State Backpropagate loss and update $ {\bm{w}} $
		\EndFor
		\label{disc_end}
		\State Sample a batch of nodes $  \bm{x}_{v} (v\in \bm{B}) $ and compute embeddings $ \bm{u}_{v} $
		\label{gen_start}
		\State Compute $ \mathcal{L}_{\rm GEN} $ using Eq.~\ref{eq_4}
		\State Backpropagate loss and update $ {\bm{\varphi}} $
		\label{gen_end}
		\EndFor
	\end{algorithmic}	
\end{algorithm}
The minibatch training procedure of AGAIN is shown in Algorithm~\ref{alg_1}. In the inductive learning phase, GNN encoder $ f_{\bm{\varphi}}(\cdot) $ and node classifier $ l_{\bm{\psi}}(\cdot) $ are updated to minimize the cross-entropy loss of labeled nodes (Steps~\ref{emb_start}--~\ref{emb_end}). The total labeled nodes of an attributed graph are randomly shuffled first, and then equally divided into several batches which are then processed one by one. Therefore, a batch of labeled nodes can be considered to be randomly sampled from the total labeled nodes. Taking one of the selected labeled nodes as a target node, we sample its neighboring nodes and aggregate neighborhood information to compute its embedding vector and label prediction. Cross-entropy loss is calculated based on the predictions and ground-truth labels using Eq.~\ref{eq_1}. When executing from Steps~\ref{emb_start}--~\ref{emb_end}, although only one batch of labeled nodes is considered, the whole graph, except for the test data, is accessible in the neighborhood aggregation process. In other words, in addition to the feature vector of a target node, the features of its neighboring nodes, which are sampled from the whole graph, are also involved in the computation of target node embedding vector. 

In adversarial training phase, the adversarial networks first update discriminator $ d_{\bm{w}}(\cdot) $ to tell apart real samples (vectors from prior distribution) from fake samples, i.e., embedding vectors (Steps~\ref{disc_start}--~\ref{disc_end}). In addition, GNN encoder, $ f_{\bm{\varphi}}(\cdot) $, serve as a generator to confuse the trained discriminator and update itself (Steps~\ref{gen_start}--~\ref{gen_end}). Therefore, the parameters of GNN encoder $ f_{\bm{\varphi}}(\cdot) $ are updated by inductive learning and adversarial training alternatively.

The computational complexity of inductive learning is proportional to the number of parameters $ \left| \bm{\varphi} \right| $ and $ \left| \bm{\psi} \right| $ in every epoch. Hence, the overall complexity is $ O(n_{\rm max}(\left| \bm{\varphi} \right| + \left| \bm{\psi} \right|)) $. Similarly, the computational complexity of generator and discriminator is typically linear with the number of parameters $ \left| \bm{\varphi} \right| $ and $ \left| \bm{w} \right| $, respectively. Therefore, the complexity of adversarial training is $ O(n_{\rm max}(n_{\rm D}\left| \bm{w} \right| + \left| \bm{\varphi} \right|)) $.
\begin{algorithm}[tbp]
	\caption{AGAIN Testing}
	\label{alg_test}
	\begin{algorithmic}[1]
		\Require Graph $ \mathcal{G}\left( \bm{V},\bm{E},\bm{X}\right)  $; test nodes (i.e., $ \bm{V}_{\rm U}^{\rm n} $); maximum search depth $ K $; trained GNN encoder $ f_{\bm{\varphi}}\left( \cdot\right)  $ (including attention-based aggregator function $ {\rm AGG}_{k} $,  weight matrices $ \bm{W}_{v}^{k} $ and $ \bm{W}_{S}^{k} $, and nonlinearity $ \sigma_{2} $, $ k\in\left\lbrace 1,\ldots,K \right\rbrace$); trained node classifier $ l_{\bm{\psi}}\left( \cdot\right) $.					 
		\State Set the initial representations as $ \bm{h}_{v}^{0}=\bm{x}_{v} (v\in \bm{V}_{\rm U}^{\rm n}) $ and sample the neighboring features $ \bm{x}_{S} $ (including those of the immediate neighbors, i.e., $ \bm{S}_{v} $).\label{emb_start_test}			
		\For {$ k=1,\ldots,K $}		 
		\State $ \bm{h}_{S}^{k}={\rm AGG}_{k}(\bm{h}_{u}^{k-1}\mid u\in \bm{S}_{v}) $
		\State $ \bm{h}_{v}^{k}=\sigma_{2}({[\bm{W}_{v}^{k}\bm{h}_{v}^{k-1}; \bm{W}_{S}^{k}\bm{h}_{S}^{k}]}) $
		\State $ \bm{h}_{v}^{k}=\bm{h}_{v}^{k}/\parallel \bm{h}_{v}^{k} \parallel_{2}$
		\EndFor
		\State $ \bm{u}_{v}=\bm{h}_{v}^{K}=f_{\bm{\varphi}}(\bm{x}_{v}, \bm{x}_{S}) $, $ \hat{\bm{y}}_{v}=l_{\bm{\psi}}(\bm{u}_{v}) $.
		\label{emb_end_test}
		\State Calculate the classification accuracy based on label prediction $ \hat{\bm{y}}_{v} $ and ground truth label.			
	\end{algorithmic}	
\end{algorithm}
\subsubsection{AGAIN Testing}
Algorithm~\ref{alg_test} outlines the process of testing. In the testing phase, since GNN encoder $ f_{\bm{\varphi}}\left( \cdot\right) $ and node classifier $ l_{\bm{\psi}}\left( \cdot\right)  $ have been trained, their learnable parameters are fixed. Then the features of test nodes and their sampled neighboring features are fed into the trained model to obtain node embeddings and label predictions (Steps~\ref{emb_start_test}--~\ref{emb_end_test}). Finally, the classification accuracy is calculated based on the label predictions and ground truth labels.

It can be seen in \textbf{Figure~\ref{fig_3_1}} that, in the test phase, the local neighborhoods of existing nodes would change, and the local structures of new nodes are newly formed. The proposed inductive learning model can handle both situations and compute the representations for all nodes in the graph. GNN encoder, $ f_{\bm{\varphi}}\left( \cdot\right) $, generates node representations by aggregating the neighborhood information (see Eq.~\ref{eq_0}--Eq.~\ref{eq_emb_k_2}). Once the GNN encoder is trained based on the available information of the original graph, its parameters are fixed in the test phase. Although the graph topology has changed during testing, the GNN encoder can still compute node representations. In Eq.~\ref{eq_0}, as long as the updated or new neighborhood information is provided, the representation of neighborhood can be computed, enabling subsequent calculation.
\section{Experiments}
\label{exp}
In this section, we aim to answer the following research questions (RQs) by extensive experiments.
\begin{itemize}	
	\item RQ1: How does AGAIN perform on the inductive node classification tasks compared with the state-of-the-art baselines?
	\item RQ2: What are the benefits of learning strategies, including information aggregation, attention mechanism, and adversarial training?
	\item RQ3: How is the performance of AGAIN model affected by the relevant hyperparameters?
\end{itemize}
\subsection{Experimental Setup}
\label{exp_set}
\paragraph{Datasets} We conduct experiments on four real-world datasets as described in \textbf{Table~\ref{tab_1}}. The three citation graphs \cite{sen_collective_2008} (i.e., Cora, CiteSeer and PubMed) have nodes and edges representing publications and citation links, respectively. These publications are categorized based on their corresponding research topics. For example, Cora consists of machine learning papers which belong to one of the seven classes named as ``case based'', ``genetic algorithms", ``neural networks", ``probabilistic methods", ``reinforcement learning", ``rule learning", and ``theory". For Cora and CiteSeer, each paper is described by a feature vector with binary values indicating whether each word from a dictionary is present. The publications in PubMed have features described by Term Frequency–Inverse Document Frequency (TF-IDF) vectors drawn from a dictionary consisting of 500 unique words. Therefore, for each citation network, the feature dimension of a node, $ D $, is determined by the corresponding dictionary size.  

BlogCatalog \cite{huang_label_2017} is an online community in which bloggers follow each other. It is modeled as a social network, with nodes and edges representing bloggers and their following relationships, respectively. The feature vector of each blogger is obtained according to the corresponding blog description. The bloggers are categorized into one of the six predefined categories based on their interests.

\begin{table}[tbp]
	\caption{Summary of datasets.}
	\label{tab_1}
	\centering
	\centerline{
	\begin{threeparttable}
	\begin{tabular}{lccccc}
		\toprule
		Dataset     & \#Nodes\tnote{$ \ast $} $ N $  & \#Edges $ \left| \bm{E} \right| $ & Average Degree $ \left\langle k \right\rangle $ & \#Labels $ C $ & \#Features $ D $ \\ \midrule
		Cora        & 2,708  &           5,429           &               4.0                &   7   & 1,433 \\
		CiteSeer    & 3,327  &           4,732           &               2.8                &   6   & 3,703 \\
		PubMed      & 19,717 &          44,338           &               4.5                &   3   &  500  \\ \midrule
		BlogCatalog & 5,196  &          171,743          &               66.1               &   6   & 8,189 \\ \bottomrule
	\end{tabular}
\begin{tablenotes}
	\footnotesize
	\item[$ \ast $] ``\#Nodes" means the number of nodes. The rest can be deduced by analogy.
\end{tablenotes}
\end{threeparttable}}
\end{table}

The goal of node classification in this work is to classify one publication into a certain research topic, or predict the interest of a blogger. Note that we treat all networks here as undirected graphs. In the performance study (Section~\ref{per_sty}), the labeled nodes of each citation graph are the same as the designated ones in the Planetoid paper \cite{yang_revisiting_2016} for a fair comparison. In the remaining experiments of Section~\ref{exp}, the labeled nodes are randomly selected from the training data. Specifically, we randomly choose the same number of labeled nodes for each class in the training nodes. In \cite{yang_revisiting_2016}, the number of labeled nodes per class (i.e., $ n $) is fixed as 20. However, in this work, labeled number, $ n $, varies from 20 to 100 for a more thorough investigation. The remaining training nodes are unlabeled. Under inductive setting, the test nodes are unobserved during training. Following the setting of \cite{yang_revisiting_2016}, the number of test nodes in each graph is fixed as 1000.

\paragraph{Baselines}
Three groups of baselines are introduced as follows.

\begin{itemize}	
	\item LR, DeepWalk~\cite{perozzi_deepwalk_2014}, and DeepWalk+: They are unsupervised baselines followed by the logistic regression classifier. LR is directly trained on the node features. DeepWalk generates embedding vector for each node using the graph structure only. In DeepWalk+, node embeddings generated by DeepWalk are further concatenated with node features. 
	\item ManiReg~\cite{belkin_manifold_2006}, SemiEmb~\cite{weston_deep_2008}, and Planetoid-I~\cite{yang_revisiting_2016}: They are graph semi-supervised learning methods. Graph Laplacian regularization is employed in these methods to impose penalty, if nearby nodes are predicted to have different labels. They are inductive baselines which can naturally handle unseen nodes. 
	\item GAT~\cite{velickovic_graph_2018} and GraphSAGE~\cite{hamilton_inductive_2017}: They are GNN models for inductive learning on graphs. GAT devises an attention mechanism to assign learnable weights for the entire neighborhood nodes. The GraphSAGE variants (including GS-GCN, GS-mean, GS-LSTM and GS-pool) employ various aggregator functions to aggregate information from the sampled neighborhood. Among them, GS-GCN is the inductive variant of the GCN model~\cite{kipf_semi-supervised_2017}. GS-mean improves on GS-GCN by concatenating the output of previous layer with a skip connection. Such skip connection can also be found in GS-LSTM and GS-pool.	
\end{itemize}

\paragraph{Implementation Details}
For baselines using logistic regression (i.e., LR, DeepWalk and DeepWalk+), we use the logistic SGDClassifier in the scikit-learn Python package~\cite{pedregosa_scikit-learn_2011} with default settings. For DeepWalk, we follow what is done in GraphSAGE~\cite{hamilton_inductive_2017}. While fixing the embeddings of already trained nodes, before making predictions, a new round of SGD optimization is performed to update the embeddings of new test nodes. For Planetoid-I, we use the public source code\footnote{\url{https://github.com/kimiyoung/planetoid}} provided by the authors with default settings, and sweep learning rate in the set $ \left\lbrace0.1, 0.01, 0.001\right\rbrace $. 

The PyTorch implementation\footnote{\url{https://github.com/Diego999/pyGAT}} of GAT model is originally transductive. We adapt this implementation to calculate the GAT results under inductive scenario. The original GraphSAGE variants only have unsupervised and fully-supervised versions. We adapt the fully-supervised version to be semi-supervised, which only has a few labeled nodes during training. Since the graph structure has already been incorporated in the neighborhood sampling process, similar to GCN~\cite{kipf_semi-supervised_2017}, GraphSAGE variants are directly trained on the supervised loss of labeled nodes, without having to consider the Laplacian regularization.

\begin{figure}[tbp]
	\centering
	\includegraphics[width=7.78cm]{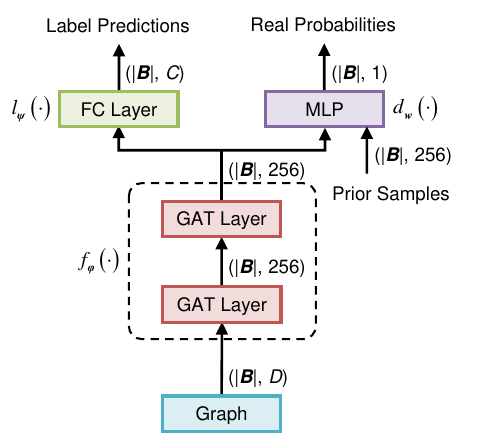}
	\caption{Layer structure of AGAIN. The input and output dimensions of each block are shown in parentheses, where $ \left| \bm{B} \right| $ is the batchsize, $ D $ is the feature dimension of a graph, and $ C $ is the number of classes.}
	\label{fig_imple}
\end{figure}

\textbf{Figure~\ref{fig_imple}} shows the layer structure of AGAIN. GNN encoder, $ f_{\bm{\varphi}}(\cdot) $, is a two-layer graph attention model. The dimension of weighting vector, $ \bm{a} $, is selected in the set $ \left\lbrace 64, 128, 256, 512, 1024, 2048\right\rbrace $. The output dimension of each layer (i.e., $ \left| \bm{h}_{v}^{k}\right|  $ in Algorithm~\ref{alg_1}) is set as 256. Node classifier, $ l_{\bm{\psi}}(\cdot) $, is a fully-connected single-layer neural network (i.e., FC layer) followed by a softmax activation. Its input and output dimensions are embedding dimension (i.e., $ d=256 $) and the number of classes in each dataset (i.e., $ C $), respectively. Discriminator, $ d_{\bm{w}}(\cdot) $, is a four-layer neural network (i.e., MLP), with the dimensions of three hidden layers set as 1024, 1024 and 256 in sequence. The output of discriminator is of one dimension, indicating the probability of an input sample to be real. We use the leaky ReLU activation (i.e., $ \sigma(x)={\rm max}(0.2x,x) $) in the first three layers, and employ a sigmoid activation in the output layer. The default prior of AGAIN is a multivariate Gaussian distribution $ \mathcal N(\pmb{0},10^{p}\pmb{I}) $. The dimension of a prior sample is the same as embedding dimension $ d $. Power exponent, $ p $, is swept in the set $ \left\lbrace -4, -2, 0, 2, 4\right\rbrace  $.

\begin{table}[tbp]
	\setlength\tabcolsep{2pt}
	\caption{Main hyperparameters for the AGAIN model.}
	\label{tab_hyper}
	\centering
	\centerline{
	\begin{threeparttable}		
	\begin{tabular}{c|c|c|c|c|c|c|c|c|c|c|c|c}
		\hline
		  \multirow{2}*{Dataset}   & \multirow{2}*{$ n $} & \multirow{2}*{$ n_{\rm max} $} & \multirow{2}*{$ n_{\rm D} $} & \multirow{2}*{Batchsize}       &         \multicolumn{2}{c|}{Learning Rate}          &        \multirow{2}*{Weight Decay}  &   \multirow{2}*{$ d $}   &  \multirow{2}*{$ p $} & \multirow{2}*{Dropout} & \multirow{2}*{$ K $} & \multirow{2}*{$ \bm{s} $\tnote{$ \ast $}}\\ \cline{6-7}
		                           &                      &                                &                              &                                                                        & $ \bm{\varphi}, \bm{\psi} $ &      $ \bm{w} $       &  &         & &                        &        &                  \\ \hline
		\multirow{3}*{BlogCatalog} &          20          &      \multirow{12}*{200}       &       \multirow{3}*{5}       &  \multirow{12}*{256}    &    \multirow{12}*{0.001}    & \multirow{3}*{0.0002} &    \multirow{3}*{0.005}     & \multirow{12}*{256} &  \multirow{3}*{0} &  \multirow{12}*{0.5}   &   \multirow{12}*{2} & \multirow{12}*{$\left\lbrace25,10\right\rbrace$}   \\ \cline{2-2}
		                           &          60          &                                &                              &                                                                        &                             &                       &                             &                   &   &                        &            &              \\ \cline{2-2}
		                           &         100          &                                &                              &                                                                        &                             &                       &                             &                   & &                        &      &                    \\ \cline{1-2}\cline{4-4}\cline{7-8}\cline{10-10}
		   \multirow{3}*{Cora}     &          20          &                                &       \multirow{9}*{1}       &                                                                        &                             &        0.0001         &     \multirow{9}*{0.05}     &                   & \multirow{9}*{$ -4 $} &                        &             &             \\ \cline{2-2}\cline{7-7}
		                           &          60          &                                &                              &                                                                        &                             &         0.001         &                             &                   &  &                        &             &             \\ \cline{2-2}\cline{7-7}
		                           &         100          &                                &                              &                                                                        &                             &        0.0001         &                             &                   & &                        &         &                 \\ \cline{1-2}\cline{7-7}
		 \multirow{3}*{CiteSeer}   &          20          &                                &                              &                                                                        &                             & \multirow{2}*{0.001}  &                             &                   & &                        &             &             \\ \cline{2-2}
		                           &          60          &                                &                              &                                                                        &                             &                       &                             &                   & &                        &      &                    \\ \cline{2-2}\cline{7-7}
		                           &         100          &                                &                              &                                                                        &                             &        0.0001         &                             &                   & &                        &        &                  \\ \cline{1-2}\cline{7-7}
		  \multirow{3}*{PubMed}    &          20          &                                &                              &                                                                        &                             & \multirow{3}*{0.001}  &                             &                   & &                        &          &                \\ \cline{2-2}
		                           &          60          &                                &                              &                                                                        &                             &                       &                             &                   & &                        &             &             \\ \cline{2-2}
		                           &         100          &                                &                              &                                                                        &                             &                       &                             &                   & &                        &               &           \\ \hline
	\end{tabular}
    \begin{tablenotes}
    	\footnotesize
    	\item[$ \ast $] Neighborhood sample size, $ \bm{s} $, is denoted as a set $ \left\lbrace s_{1}, \ldots, s_{K}\right\rbrace  $ containing sample size $ s_{k} $ in each search depth $ k $.
   \end{tablenotes}	    	
\end{threeparttable}}
\end{table}

AGAIN is implemented using PyTorch~\cite{paszke_pytorch_2019}. In \textbf{Table~\ref{tab_hyper}}, we provide the main hyperparameters selected for each dataset, with $ n $ denoting the number of labeled training nodes per class. We train the model for 200 epochs (i.e., $ n_{\rm max}=200 $) using the Adam optimizer. The batch size (i.e., $ \left| \bm{B} \right| $) is 256. In order to prevent overfitting, $ L_{2} $ regularization is enforced in the loss function with the weight decay term selected in $ \left\lbrace 5e-5, 5e-4, 5e-3, 5e-2 \right\rbrace $. All weights are initialized by default. Note that we set the maximum search depth as $ K=2 $, in which neighborhood sample sizes are $ s_{1}=25 $ and $ s_{2}=10 $, respectively. As mentioned in \cite{hamilton_inductive_2017}, increasing $ K $ beyond 2 leads to marginal accuracy improvement, while large increment can be seen in the corresponding runtime. The influence of sample size (i.e., $ s $) on classification accuracy is discussed in Section~\ref{para_sty}.

For fair comparison, the above methods have the same embedding dimension, i.e., $ d=256 $. All results are averaged by ten runs with different random seeds. We run the experiments using a computer with one NVIDIA GeForce GTX 1080Ti GPU (11 GB of RAM), an Intel(R) Core(TM) i7-8700K CPU (6 cores, 3.70 GHz), and 32 GB of RAM.
\subsection{Performance Study (RQ1)}
\label{per_sty}
In this section, experiments are first conducted on the inductive benchmark task to verify the proposed method. Then the t-SNE visualization of node representation is provided. Finally, the proposed methods are further evaluated under the transductive setting. 
\subsubsection{Inductive Node Classification}
Following the dataset split of the Planetoid paper~\cite{yang_revisiting_2016}, in the training nodes, there are only 20 labeled nodes for each class, i.e., $ n=20 $. 1000 nodes are selected as the test data. In \textbf{Table~\ref{tab_2}}, for the citation datasets, the accuracies of ManiReg, SemiEmb, and Planetoid-I are taken from \cite{yang_revisiting_2016}. Since the results of these methods on BlogCatalog are not reported in \cite{yang_revisiting_2016}, the corresponding cells of ManiReg and SemiEmb are left empty. As stated in Section~\ref{exp_set}, the performance of Planetoid-I in BlogCatalog is obtained by executing the source code. In the subsequent discussions, AGAIN is the full proposed model shown in \textbf{Figure~\ref{fig_1}}. GAIN is our simplified model without adversarial training. That is, GAIN only consists of the GNN encoder and the node classifier. It can be seen in \textbf{Table~\ref{tab_2}} that, logistic regression classifier (LR) produces the largest standard deviation values, possibly due to its simplicity. When testing on the same graph, the standard deviation values of other methods are generally close and of the same order. Therefore, the following discussions are based on the mean accuracy. 
\begin{table}[tbp]
	\caption{Mean classification accuracy on test data under inductive setting (in percent). For each dataset, the highest mean accuracy is highlighted in bold and the top two are underlined. The standard deviations are given in parentheses.}
	\label{tab_2}
	\centering
	\begin{tabular}{lcccc}
		\toprule
		Method                                  &              BlogCatalog              &                 Cora                  &               CiteSeer                &                PubMed                 \\ \midrule
		LR                                      &           66.4 (3.6)             &           51.6 (2.6)            &           51.0 (1.5)            &           71.4 (3.5)            \\
		DeepWalk~\cite{perozzi_deepwalk_2014}   &           25.4 (1.3)            &           29.4 (1.3)            &           22.9 (1.1)            &           48.2 (2.2)            \\
		DeepWalk+                               &           66.5 (2.5)            &           55.9 (0.9)            &           49.0 (0.6)            &           67.2 (0.8)            \\ \midrule
		ManiReg~\cite{belkin_manifold_2006}     &                   -                   &                 59.5                  &                 60.1                  &                 70.7                  \\
		SemiEmb~\cite{weston_deep_2008}         &                   -                   &                 59.0                  &                 59.6                  &                 71.1                  \\
		Planetoid-I~\cite{yang_revisiting_2016} &           73.2 (2.0)            &                 61.2                  &                 64.7                  &           77.2            \\ \midrule
		GAT~\cite{velickovic_graph_2018}        &           63.7 (2.6)            & \underline{\textbf{80.6}} (0.4) &           67.7 (1.0)            & \underline{\textbf{77.8}} (0.7) \\
		GS-GCN~\cite{hamilton_inductive_2017}   &           59.2 (2.7)            &           77.6 (1.2)            &           67.4 (0.5)            &           76.0 (0.7)            \\
		GS-mean~\cite{hamilton_inductive_2017}  &           77.1 (2.3)            &           79.8 (0.5)            &           68.8 (0.5)            &           76.9 (0.6)            \\
		GS-LSTM~\cite{hamilton_inductive_2017}  &           74.5 (1.9)            &           78.4 (0.4)            &           67.2 (1.1)            &           76.0 (0.7)            \\
		GS-pool~\cite{hamilton_inductive_2017}  &           73.9 (2.0)            &     \underline{80.2} (0.7)      &           68.1 (0.7)            &           77.1 (0.5)            \\ \midrule
		GAIN [ours]                             &     \underline{79.3} (1.9)      &           80.0 (0.7)            &     \underline{69.2} (0.6)      &           77.1 (0.6)            \\
		AGAIN [ours]                            & \underline{\textbf{80.1}} (1.7) &           79.9 (0.4)            & \underline{\textbf{70.0}} (0.8) & \underline{77.5} (0.7) \\ \bottomrule
	\end{tabular}
\end{table}

In the first group of baselines (i.e., LR, DeepWalk, and DeepWalk+), LR obtains much higher accuracies than DeepWalk. This indicates that, for attributed graphs, node features can be more informative than graph structure in learning node embeddings. Note that, although DeepWalk is far more competitive in transductive learning, it has poor performance on inductive tasks. Furthermore, DeepWalk performs worst in CiteSeer, which is probably attributed to the low average degree (see \textbf{Table~\ref{tab_1}}). The importance of feature information is further validated by the performance lift of DeepWalk+ compared to DeepWalk, after concatenating node features with the learned embeddings. It is found in CiteSeer and PubMed that, although utilizing both structure and feature information, DeepWalk+ cannot surpass LR. Hence, it is not always workable by simply concatenating structural embeddings and node features. In other words, graph structure and feature information need to be incorporated in a systematic manner. 

Compared with the first group of baselines, superior performance is observed in the graph-based semi-supervised learning methods (i.e., ManiReg, SemiEmb, and Planetoid-I), which is yielded by jointly incorporating the information of features, structure, and labels in an attributed graph. Among them, Planetoid-I is the most competitive one. Further improvements can be seen in the GraphSAGE variants.   

On Cora and PubMed, we observe several inductive GNN models yield close performance, including GAT, GS-mean, GS-pool, GAIN, and AGAIN. On CiteSeer and BlogCatalog, AGAIN has clear performance gains over other GNN models. Specifically, with the help of attention mechanism and skip connection, GAIN outperforms GraphSAGE variants and GAT. Then, AGAIN further improves on GAIN by adversarial training which increases the generalization ability. Compared with those of Cora and PubMed, node feature vectors have larger dimensions in BlogCatalog and CiteSeer (see \textbf{Table~\ref{tab_1}}). Thus, the above observations reveal the strength of our methods in performing inductive learning on feature-rich graphs. In particular, AGAIN outperforms GAT by 6.4\% on BlogCatalog and 2.3\% on CiteSeer. On BlogCatalog, GAT surpasses the inductive variant of GCN (i.e., GS-GCN), but underperforms other GraphSAGE variants which employ skip connection and advanced aggregator. Note that, through concatenating the output of previous layer, GS-mean outperforms GS-GCN in all cases, showing the benefit of skip connection. 
\subsubsection{Visualization of Embedding Vectors}
\begin{figure}[tbp]
	\makebox[\textwidth][c]{\includegraphics[width=15.0cm]{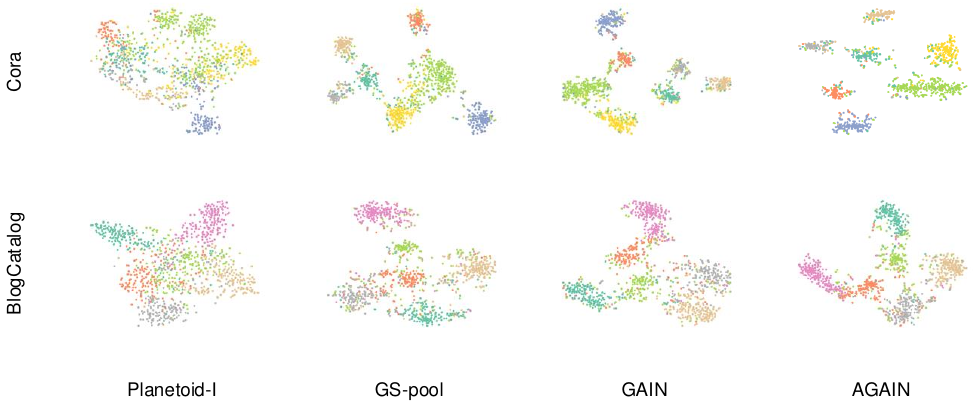}}
	\caption{Visualization of the embedding vectors of Cora and BlogCatalog in the 2D space using t-SNE (best viewed in color). For Cora, each point corresponds to one paper. Seven colors distinguish different paper classes. For BlogCatalog, each point represents one blogger. Six colors denote different interests.}
	\label{fig_2}
\end{figure}

\textbf{Figure~\ref{fig_2}} visualizes the embedding vectors of the test nodes in Cora and BlogCatalog using t-SNE \cite{maaten_visualizing_2008}. We select Planetoid-I and GS-pool as representative baselines, and neglect methods in the first group due to their low accuracies. As shown in \textbf{Table~\ref{Sil_score}}, we further calculate the corresponding Silhouette score \cite{rousseeuw_silhouettes_1987} for the clusters in a 2D projected space. Embedding vectors generated by AGAIN have the most preferable visualization. Specifically, the clusters are separated more clearly, yielding the highest Silhouette score.

\begin{table}[tbp]
	\caption{Silhouette score of the clusters in a 2D projected space.}
	\label{Sil_score}
	\centering
	\begin{tabular}{lcccc}
		\toprule
		Dataset     & Planetoid-I & GS-pool & GAIN  & AGAIN \\ \midrule
		Cora        &    0.034    &  0.298  & 0.280 & 0.325 \\
		BlogCatalog &    0.231    &  0.230  & 0.284 & 0.341 \\ \bottomrule
	\end{tabular}
\end{table}

\subsubsection{Transductive Node Classification}
\begin{table}[tbp]
	\caption{Mean classification accuracy on test data under transductive setting (in percent). For each dataset, the highest mean accuracy is highlighted in bold and the top two are underlined. The standard deviations are given in parentheses.}
	\label{tab_trans}
	\centering
	\begin{tabular}{lcccc}
		\toprule
		Method  &              BlogCatalog              &                 Cora                  &               CiteSeer                &            PubMed            \\ \midrule
		GCN~\cite{kipf_semi-supervised_2017}     &                 65.0 (2.3)                  &                 \underline{\textbf{81.5}}                  &           \underline{70.3}            &  \underline{\textbf{79.0}}   \\ \midrule
		GS-GCN~\cite{hamilton_inductive_2017}  &           59.5 (2.0)            &           78.4 (1.1)            &           67.2 (0.7)            &       76.5 (0.9)       \\
		GS-mean~\cite{hamilton_inductive_2017} &           76.2 (2.8)            &           80.0 (0.6)            &           69.2 (0.7)            &       76.6 (0.5)       \\
		GS-LSTM~\cite{hamilton_inductive_2017} &           73.7 (2.0)            &           79.4 (0.7)            &           67.4 (1.4)            &       75.6 (0.5)       \\
		GS-pool~\cite{hamilton_inductive_2017} &           73.3 (2.1)            &     80.3 (0.5)      &           68.6 (0.4)            &       77.4 (0.7)       \\ \midrule
		GAIN [ours]   &     \underline{79.2} (2.4)      &     \underline{80.4} (0.4)      &           69.6 (0.7)            &       76.6 (0.7)       \\
		AGAIN [ours]  & \underline{\textbf{79.8}} (2.2) & 80.3 (0.6) & \underline{\textbf{70.5}} (0.8) & \underline{77.6} (0.6) \\ \bottomrule
	\end{tabular}
\end{table}
Though this work aims at inductive learning, to make the evaluation more comprehensive, we further conduct experiments under the transductive settings. In \textbf{Table~\ref{tab_trans}}, the results are presented and compared with the classical tranductive GNN model, i.e., GCN~\cite{kipf_semi-supervised_2017}. We reuse the GCN results reported in~\cite{kipf_semi-supervised_2017} for Cora, CiteSeer, and PubMed. The GCN performance in BlogCatalog is evaluated by adapting and executing the source codes~\footnote{\url{https://github.com/tkipf/pygcn}} provided by the authors. Since the test set information is originally assumed to be unavailable when designing the inductive approaches, such information may not be well exploited by the inductive approaches, consequently leading to their underperformance in some cases.

It is found that, in Cora, CiteSeer and PubMed, the transductive accuracies in \textbf{Table~\ref{tab_trans}} are generally higher than the corresponding inductive ones in \textbf{Table~\ref{tab_2}}. The reason is that, unlike inductive learning, the information of test data is accessible in transductive learning during training. However, in BlogCatalog, the transductive accuracies are mostly lower than those in inductive cases. It might be attributed to the high average degree of BlogCatalog (i.e., $ 66.1 $, see \textbf{Table~\ref{tab_1}}). The sample sizes are $ 25 $ and $ 10 $ in the immediate neighbors and the 2-hop neighbors (see \textbf{Table~\ref{tab_hyper}} and \textbf{Figure~\ref{fig_3.3_1}}), respectively. Therefore, under the inductive setting, there will already be rich neighborhood information to exploit. In transductive setting, although test set information is available when sampling neighboring nodes, more abundant neighborhood information can sometimes introduce certain noise, which would lead to underperformance. 

Although AGAIN is designed for inductive learning, compared with the classical GCN model~\cite{kipf_semi-supervised_2017}, its results are still matched in CiteSeer and even higher in BlogCatalog. GCN surpasses its inductive variant GS-GCN in BlogCatalog. However, with advanced aggregator and skip connection employed, the remaining GNN models manage to yield large performance gains over GCN. Note that, in all cases, one of the proposed methods (i.e., GAIN or AGAIN) is able to outperform the GraphSAGE variants.
\subsection{Ablation Study (RQ2)}
In this section, we investigate the influences of information aggregation, attention mechanism, and adversarial training on learning node embeddings step by step. We construct a two-layer MLP, which only uses node features as input, without having to consider graph structure, and outputs predictions. The first layer of MLP is similar to GNN encoder $ f_{\bm{\varphi}}(\cdot) $ in \textbf{Figure~\ref{fig_1}}. The second layer can be treated as node classifier $ l_{\bm{\psi}}(\cdot) $. Therefore, the hidden dimension of MLP is set as embedding dimension $ d $. Referring to the neighborhood representation obtained using Eq.~\ref{eq_0}, GS-mean takes the average of the representations of neighbors with equal weights. GAIN further assigns different learnable weights (i.e., attention coefficients) to these neighboring nodes. Then AGAIN combines adversarial training with GAIN, constraining the learned embeddings to match a prior distribution. Note that the subsequent experiments are all conducted under inductive setting.

In \textbf{Table~\ref{tab_3}}, the methods based on information aggregation (i.e., GS-mean, GAIN, and AGAIN) are superior to MLP which solely exploits node features. Compared with GS-mean, GAIN obtains higher accuracies in most cases. Large margins can be seen on BlogCatalog, where GAIN achieves on average 2.93$\%$ relative gain in accuracy over GS-mean. However, the margins are small on citation graphs which are relatively sparse. This indicates that the attention mechanism can be more powerful on a dense graph containing rich features. In terms of the mean accuracy, AGAIN outperforms GAIN in 9 out of 12 cases, showing AGAIN has a slightly improved generalization ability when evaluated on unseen test nodes.
\begin{table}[tbp]
	\setlength\tabcolsep{0.5pt}
	\caption{Summary of mean classification accuracy (in percent). The number of labeled nodes per class, $ n $, varies from 20 to 100. Bold font denotes the top model. The standard deviations are given in parentheses.}
	\label{tab_3}
	\centering
	\centerline{
	\begin{tabular}{l|ccc|ccc|ccc|ccc}
		\toprule
		Dataset & \multicolumn{3}{|c|}{BlogCatalog}             & \multicolumn{3}{|c|}{Cora}                    & \multicolumn{3}{|c}{CiteSeer}                 & \multicolumn{3}{|c}{PubMed}                   \\ \midrule
		$ n $   &      20       &      60       &      100      &      20       &      60       &      100      &      20       &      60       &      100      &      20       &      60       &      100      \\ \midrule
		MLP     &  73.1 (1.6)   &  84.1 (0.9)   &  87.6 (0.5)   &  59.3 (1.2)   &  67.9 (1.2)   &  71.5 (1.3)   &  56.0 (1.2)   &  66.6 (1.8)   &  69.5 (2.0)   &  73.4 (0.5)   &  73.9 (1.9)   &  76.3 (1.6)   \\
		GS-mean &     77.1 (2.3)     &     86.1 (1.7)     &     89.2 (0.8)     &     79.8 (0.5)     &     80.9 (0.8)     &     83.2 (0.4)     &     68.8 (0.5)     &     73.0 (1.1)     &     74.2 (1.0)     &     76.9 (0.6)     &     78.3 (1.7)     &     81.4 (1.0)     \\ \midrule
		GAIN    &     79.3 (1.9)     &     89.1 (1.4)     &     91.4 (0.7)     & \textbf{80.0} (0.7) &     81.1 (1.2)     & \textbf{83.4} (1.0) &     69.2 (0.6)     & \textbf{73.6} (1.3) &     74.6 (0.9)     &     77.1 (0.6)     &     78.9 (1.6)     &     81.2 (1.1)     \\
		AGAIN   & \textbf{80.1} (1.7) & \textbf{89.2} (1.0) & \textbf{91.5} (0.8) &     79.9 (0.4)     & \textbf{82.2} (1.4) &     83.1 (1.0)     & \textbf{70.0} (0.8) &     73.1 (1.0)     & \textbf{74.9} (0.5) & \textbf{77.5} (0.7) & \textbf{79.2} (1.8) & \textbf{81.8} (1.4) \\ \bottomrule
	\end{tabular}}
\end{table}

To further investigate the effects of attention mechanism and adversarial training on improving the robustness of embeddings, we corrupt the node features in test phase after the models are trained. Referring to GIB \cite{wu_graph_2020}, we randomly choose a percentage of nodes (denoted as $ \eta $), and add independent Gaussian noise $ \left( \lambda\cdot r\cdot \epsilon \right) $ to each dimension of their feature vectors, with increasing amplitude. Random number, $ \epsilon $, is from standard normal distribution  $ \mathcal N(0,1) $. Feature noise ratio, $ \lambda $, is selected in the set $ \left\lbrace 0, 0.5, 1.0, 1.5 \right\rbrace $. When $ \lambda $ equals 0, it is the case without noise. To incorporate the graph property during noise injection, reference amplitude, $ r $, is obtained by taking the average of the maximum value in each node's feature vector. As stated in Section~\ref{rel_work}, the noise here is similar to the evasion attacks. 

In real applications, it would be a frequently encountered situation that a small fraction of nodes are noisy. Therefore, the percentage of nodes with noise, $ \eta $, is first fixed as 10$\%$. The experimental results are reported in \textbf{Table~\ref{tab_4}}. GAIN outperforms GS-mean in most cases, which indicates the potential of attention mechanism on improving robustness. Moreover, the accuracy of AGAIN is higher than that of GAIN, except those on CiteSeer. This reveals that adversarial training contributes to generating robust embeddings in some degree. For comprehensive evaluation, we also present the results of Planetoid-I and GS-pool, which are found to be mostly inferior to those of our methods. There is one exception on BlogCatalog, where Planetoid-I performs best when the noise amplitude is high (i.e., $ \lambda=1.5 $). The likely reason is that the methods relying on information aggregation are influenced by the relational effect of graph structure \cite{zugner_adversarial_2018}. Referring to \textbf{Figure~\ref{fig_3.3_1}}, the noise added on one node might misguide the predictions of other nodes with structural relations, worsening the model performance, especially when the high-intensity noise is injected into a dense graph.

\begin{table}[tbp]
	\setlength\tabcolsep{2pt}
	\caption{Mean classification accuracy in percent for the trained models with increasing additive feature noise ($ n=60, \eta=10\% $). Bold font denotes the top model.}
	\label{tab_4}
	\centering
	\centerline{
	\begin{tabular}{l|cccc|cccc|cccc|cccc}
		\toprule
		Dataset     & \multicolumn{4}{|c|}{BlogCatalog}                             & \multicolumn{4}{|c|}{Cora}                                    & \multicolumn{4}{|c}{CiteSeer}                                 & \multicolumn{4}{|c}{PubMed}                                   \\ \midrule
		$ \lambda $ &       0       &      0.5      &      1.0      &      1.5      &       0       &      0.5      &      1.0      &      1.5      &       0       &      0.5      &      1.0      &      1.5      &       0       &      0.5      &      1.0      &      1.5      \\ \midrule
		Planetoid-I &     83.0      &     77.6      &     76.9      & \textbf{76.7} &     69.2      &     66.1      &     65.0      &     64.7      &     69.3      &     66.7      &     65.4      &     65.0      &     74.7      &     71.6      &     71.0      &     70.6      \\
		GS-pool     &     84.8      &     74.2      &     63.0      &     54.6      &     80.8      &     76.3      &     67.5      &     61.2      &     72.5      &     68.4      &     62.3      &     58.4      &     78.0      &     72.7      &     69.5      &     67.6      \\
		GS-mean     &     86.1      &     80.4      &     77.7      &     74.9      &     80.9      &     77.6      &     72.4      &     68.9      &     73.0      &     70.5      &     67.3      &     64.7      &     78.3      &     74.4      &     71.5      &     69.7      \\ \midrule
		GAIN        &     89.1      &     82.0      &     75.3      &     68.4      &     81.1      &     77.9      &     72.9      &     69.7      & \textbf{73.6} & \textbf{72.0} & \textbf{68.6} & \textbf{65.9} &     78.9      &     74.8      &     71.8      &     69.7      \\
		AGAIN       & \textbf{89.2} & \textbf{82.2} & \textbf{78.2} &     72.5      & \textbf{82.2} & \textbf{78.8} & \textbf{74.7} & \textbf{71.1} &     73.1      &     71.1      &     67.7      &     65.3      & \textbf{79.2} & \textbf{75.4} & \textbf{72.7} & \textbf{71.4} \\ \bottomrule
	\end{tabular}}
\end{table}

\begin{figure}[tbp]
	\centering
	\makebox[\textwidth][c]{\includegraphics[width=18cm]{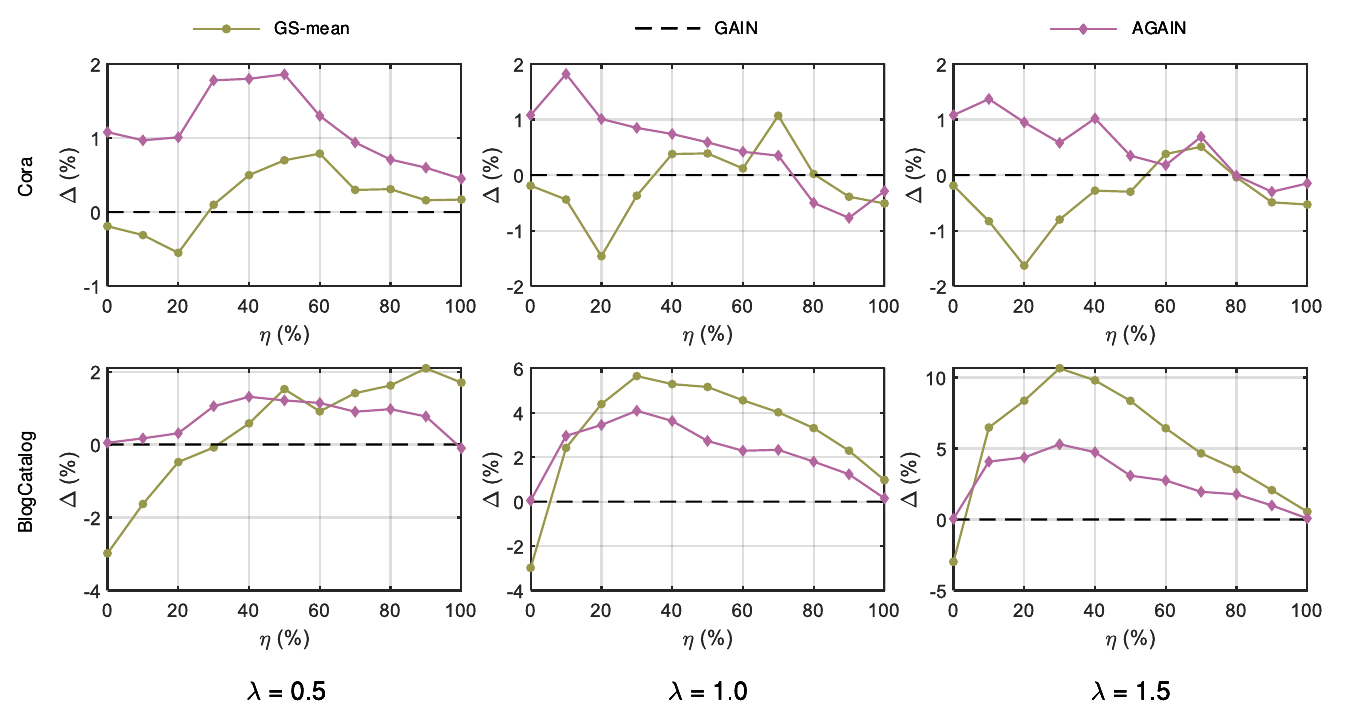}}
	\caption{Performance gap (i.e., $ \Delta $) on Cora and BlogCatalog ($ n=60 $). $ \eta $ is the percentage of nodes in a graph corrupted with additive feature noise. $ \lambda $ is the feature noise ratio.}
	\label{fig_rel_acc}
\end{figure}

Further investigations are conducted by varying the percentage of nodes with noise (i.e., $ \eta $). For the sake of clarity, performance gap, $ \Delta $, is obtained through subtracting the accuracy of GAIN from that of GS-mean or AGAIN. As shown in \textbf{Figure~\ref{fig_rel_acc}}, the performance of attention mechanism and adversarial training varies with node percentage and graph property. In general, as the percentage of nodes with noise increases, the attention mechanism brings an improvement first but loses effects gradually. With adversarial training employed, AGAIN outperforms GAIN by clear margins in most cases, revealing adversarial training is able to improve model robustness. However, the performance gain yielded by adversarial training shrinks or even becomes negative when more nodes are corrupted with noise. 
\subsection{Hyperparameter Sensitivity Study (RQ3)}
\label{para_sty}
\begin{figure}[tbp]
	\centering
	\makebox[\textwidth][c]{\includegraphics[width=15.3cm]{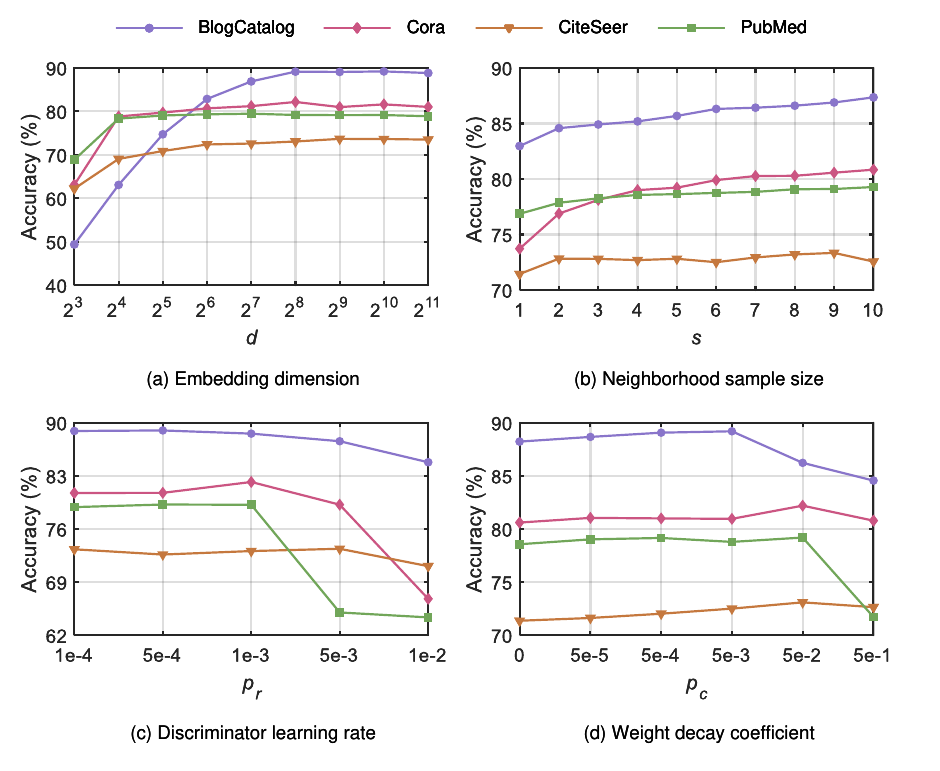}}
	\caption{Accuracy varied with four different hyperparameters individually ($ n=60 $).}
	\label{fig_3}
\end{figure}
In this section, we analyze the classification accuracy of AGAIN with regard to four relevant hyperparameters, i.e., embedding dimension $ d $, neighborhood sample size $ s $, discriminator learning rate $ p_{r} $, and weight decay coefficient $ p_{c} $. When one hyperparameter is investigated, the remaining hyperparameters are set as the default values introduced in Section~\ref{exp_set}. \textbf{Figure~\ref{fig_3}} displays the classification accuracies on the four graphs.
	
Embedding dimension, $ d $, is the dimension of node representation vector learned by the AGAIN model. The prediction accuracy increases with the embedding dimension first and then becomes stable. Similar trends can be seen on Cora and PubMed, when increasing the number of sampled neighbors (i.e., $ s $). However, there are little variations on CiteSeer due to its low average degree (see \textbf{Table~\ref{tab_1}}). In contrast, when the test is applied on BlogCatalog which has relatively high density, the accuracy increases steadily with the sample size. Note that, when investigating the sample size, we select the same number of neighbors in each search depth, i.e., $ s_{1}=s_{2}=s $ (see \textbf{Table~\ref{tab_hyper}}). In the case that the learning rate of discriminator (i.e., 1e-2) is much larger than that of the GNN encoder (i.e., 1e-3), a clear performance drop is observed on each graph. When evaluated on Cora and PubMed, the model is more sensitive to the discriminator learning rate. On BlogCatalog, the best accuracy is obtained with a weight decay coefficient of 5e-3. On the three citation graphs (i.e., Cora, CiteSeer, and PubMed), the classification accuracy reaches its peak value when $ p_{c}= $ 5e-2.
\section{Conclusion}
\label{con}
An adversarially regularized GNN model, AGAIN, has been proposed to address the inductive node classification problem on partially labeled graphs. AGAIN generates an informative representation vector for an unseen node with an attention-based aggregator that aggregates information from its neighbors. Adversarial training is employed to improve model robustness and generalization ability by matching node representations with a prior distribution. Experimental results on inductive node classification tasks show that our method achieves matched or even more favorable performance compared with the state-of-the-art methods.
\section*{Acknowledgments}
This work was supported in part by the Research Grants Council (RGC) of Hong Kong (17201820, 17207020, 17205919), as well as the Innovation and Technology Commission (ITC), Hong Kong (MRP/029/20X), and Centre for Transformative Garment Production (TransGP) funded by ITC.

\bibliography{mybibfile}

\end{document}